\documentclass[conference,table]{IEEEtran}
\IEEEoverridecommandlockouts
\usepackage{bbding}
\usepackage{cite}
\usepackage{amsmath,amssymb,amsfonts}
\usepackage{textcomp}
\usepackage{caption}
\usepackage{graphicx}
\usepackage{pgfplots}
\pgfplotsset{compat=1.18}
\usepackage{subcaption}
\usepackage{multirow}
\usepackage{algpseudocode}
\usepackage{algorithm}
\AtBeginDocument{%
  \providecommand\BibTeX{{%
    \normalfont B\kern-0.5em{\scshape i\kern-0.25em b}\kern-0.8em\TeX}}}

\usepackage{comment}
\usepackage{authblk}
\usepackage{amssymb}
\usepackage{fancyhdr}
\fancypagestyle{firstpage}
{
    \fancyhead[L]{\footnotesize © 2024 IEEE.  Personal use of this material is permitted.  Permission from IEEE must be obtained for all other uses, in any current or future media, including reprinting/republishing this material for advertising or promotional purposes, creating new collective works, for resale or redistribution to servers or lists, or reuse of any copyrighted component of this work in other works. This paper is accepted at the 29th IEEE European Test Symposium 2024 (ETS) 2024.}
    \fancyhead[R]{}
}
\begin{document}

\title{AdAM: Adaptive Fault-Tolerant Approximate Multiplier for Edge DNN Accelerators}
\author{
{Mahdi Taheri}\textsuperscript{1},\
{Natalia Cherezova}\textsuperscript{1}\textsuperscript{*},\
{Samira Nazari}\textsuperscript{3}\textsuperscript{*},\
{Ahsan Rafiq}\textsuperscript{1},\
{Ali Azarpeyvand}\textsuperscript{1,3},\\
{Tara Ghasempouri}\textsuperscript{1},\
{Masoud Daneshtalab}\textsuperscript{1,2},\
{Jaan Raik}}
\author[1]{ Maksim Jenihhin}

\affil[1]{Tallinn University of Technology, Tallinn, Estonia}
\affil[2]{Mälardalen University, Västerås, Sweden}
\affil[3]{University of Zanjan, Zanjan, Iran}
\affil[1]{mahdi.taheri@taltech.ee}

\maketitle
\thispagestyle{firstpage}
\begin{abstract}
Multiplication is the most resource-hungry operation in the neural network’s processing elements. In this paper, we propose an architecture of a novel adaptive fault-tolerant approximate multiplier tailored for ASIC-based DNN accelerators. AdAM employs an adaptive adder relying on an unconventional use of the leading one position value of the inputs for fault detection through the optimization of unutilized adder resources. The proposed architecture uses a lightweight fault mitigation technique that sets the detected faulty bits to zero. The hardware resource utilization and the DNN accelerator’s reliability metrics are used to compare the proposed solution against the triple modular redundancy (TMR) in multiplication, unprotected exact multiplication, and unprotected approximate multiplication. It is demonstrated that the proposed architecture enables a multiplication with a reliability level close to the multipliers protected by TMR utilizing 63.54\% less area and having 39.06\% lower power-delay product compared to the exact multiplier.

\end{abstract}
\let\thefootnote\relax\footnote{* These authors contributed equally}
\section{Introduction}

The role of Deep Neural Networks (DNNs) in a wide range of safety- and mission-critical applications (e.g., autonomous driving) is expanding. Therefore, deployment of a DNN accelerator requires addressing the trade-off between different design parameters and \textit{reliability} \cite{ahmadilivani2023systematic} \cite{taheri2022dnn}. Even though DNNs possess certain intrinsic fault-tolerant and error-resilient characteristics, it is insufficient to conclude the reliability of DNNs without considering the different characteristics of a hardware accelerator for vital applications. 
With the continuous scaling-down of the process, there is a discernible trend indicating that the Soft Error Rate (SER) of combinational circuits may surpass that of sequential circuits \cite{mahatme2011comparison} \cite{taheri2022novel}. Therefore, the main focus of this study is introducing a novel reliability technique to mitigate the soft errors in the combinational logic of an AI computation core.

This work presents the architecture of a novel adaptive fault-tolerant approximate multiplier (AdAM) tailored for ASIC-based DNN accelerators. Yet, the proposed multiplier can be implemented on FPGA as well. The contributions of the paper are as follows:
\begin{itemize}
    \item The architecture of a novel adaptive fault-tolerant approximate multiplier tailored for DNN accelerators, including an adaptive adder relying on an unconventional use of the leading one position value of the inputs for fault detection through the optimization of unutilized adder resources
    \item Implementation and validation of the multiplier in a design synthesized for ASIC
    \item Reliability assessment and comparison of the proposed multiplier with exact and approximate state-of-the-art multipliers using several DNN benchmarks
\end{itemize}
The main objective of the proposed multiplier is to have the best trade-off between power-delay product (PDP) and vulnerability (accuracy drop due to the fault) which is demonstrated in the results section. Moreover, it is demonstrated that the proposed architecture enables a multiplication with a reliability level close to the multipliers protected by TMR utilizing 63.54\% less area and having 39.06\% lower PDP compared to the exact multiplier.

The remainder of the paper is organized as follows. Section II summarizes related works, the proposed method is presented in Section III, Section IV provides the experimental setup and discusses the results, and finally, the work is concluded in Section V.


\section{Related Works}

Multipliers are one of the primary arithmetic building blocks widely used in DNNs. Various approximate multipliers are proposed in the literature. 
ScaleTRIM is a scalable approximate unsigned LOD multiplier for DNNs that exploits curve fitting and linearization for fitting input products and a novel error compensation method using lookup tables \cite{farahmand2023scaletrim}. More details about recent works on the approximation for DNNs can be found in \cite{Henkel}.

Error introduced by approximation is deterministic and its impact can be studied on the accuracy drop of the network comprehensively. However, soft errors are unpredictable effects in contaminated and harsh environments that may lead to DNNs malfunction and accuracy drop drastically \cite{Guanpeng}. Recent research investigates the reliability of DNNs alongside approximation \cite{10140043}\cite{taheri2023appraiser} and quantization \cite{taheri2024exploration}. In \cite{taheri2023deepaxe}, DNNs and approximated DNNs are tested in the presence of faults, and the results demonstrated that approximated DNNs are more resilient under special conditions.


To increase reliability and mitigate faults, Triple Modular Redundancy (TMR) and Gate-Sizing (GS) are two well-established hardening methods widely employed to mitigate the soft error rate in combinational circuits. Despite achieving 100\% fault coverage for a single fault in one module of a combinational circuit, TMR incurs a substantial near 200\% area and power overhead \cite{mittal2020survey}. Therefore, numerous algorithms and frameworks are developed to enhance the efficiency of applying these methods and balance their hardening effects and design costs 
\cite{ammes2023atmr}.

 Approximate TMR (ATMR) is a technique that replaces some modules of TMR with approximate ones while ensuring the majority voter gives the correct output \cite{arifeen2020approximate}. However, ATMR still requires duplicating the whole combinational circuit, even at the finest level of granularity.

To tackle this issue, this work presents an adaptive reliable multiplier that provides a high level of reliability while using less area than an exact multiplier.

\section{AdAM architecture}
The proposed architecture is an adaptive fault-tolerant approximate multiplier tailored for DNN accelerators. This architecture includes an adaptive adder relying on an unconventional use of the leading one position value of the inputs for fault \textit{detection} and \textit{mitigation} through the optimization of unutilized adder resources.
The proposed multiplier is an adaptation of the classical Mitchell multiplier \cite{Mitchell}. Mitchell multiplier employs approximate logarithms of the input values. By adding these logarithms, Mitchell’s algorithm estimates the product. The final result is obtained by taking the antilogarithm of this sum. Another level of approximation is introduced in the adaptive adder considering the application of this multiplier in DNNs with a proven negligible impact on the network accuracy (see Subsection \ref{adapt_add} and Table \ref{acc}).


\begin{figure}[h]
    \centering
    \includegraphics[width = 0.25\textwidth]{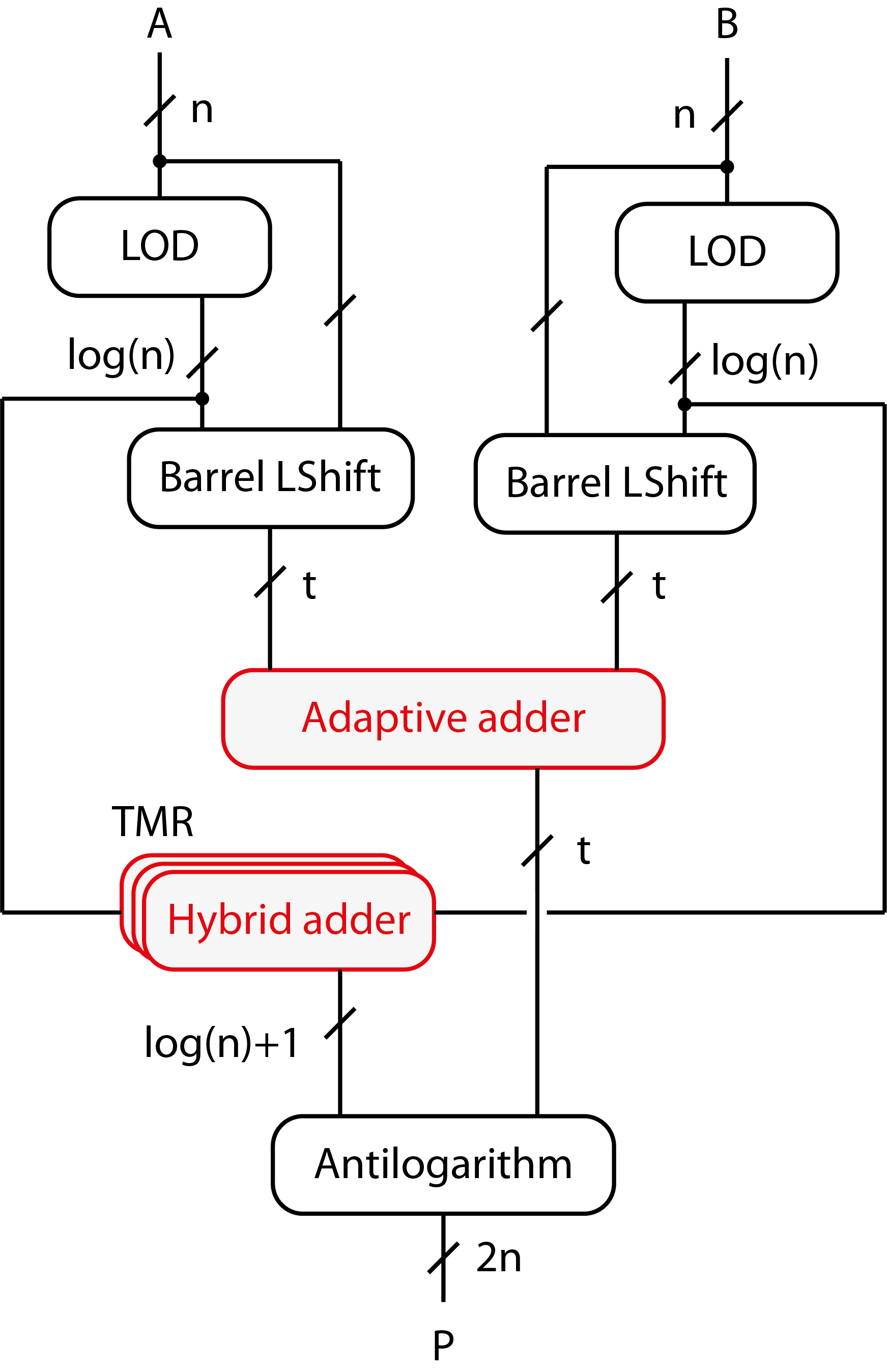}
    \caption{AdAM architecture}
    \label{fig:mult}
\end{figure}

\begin{figure}[h]
    \centering
    \includegraphics[width = 0.25\textwidth]{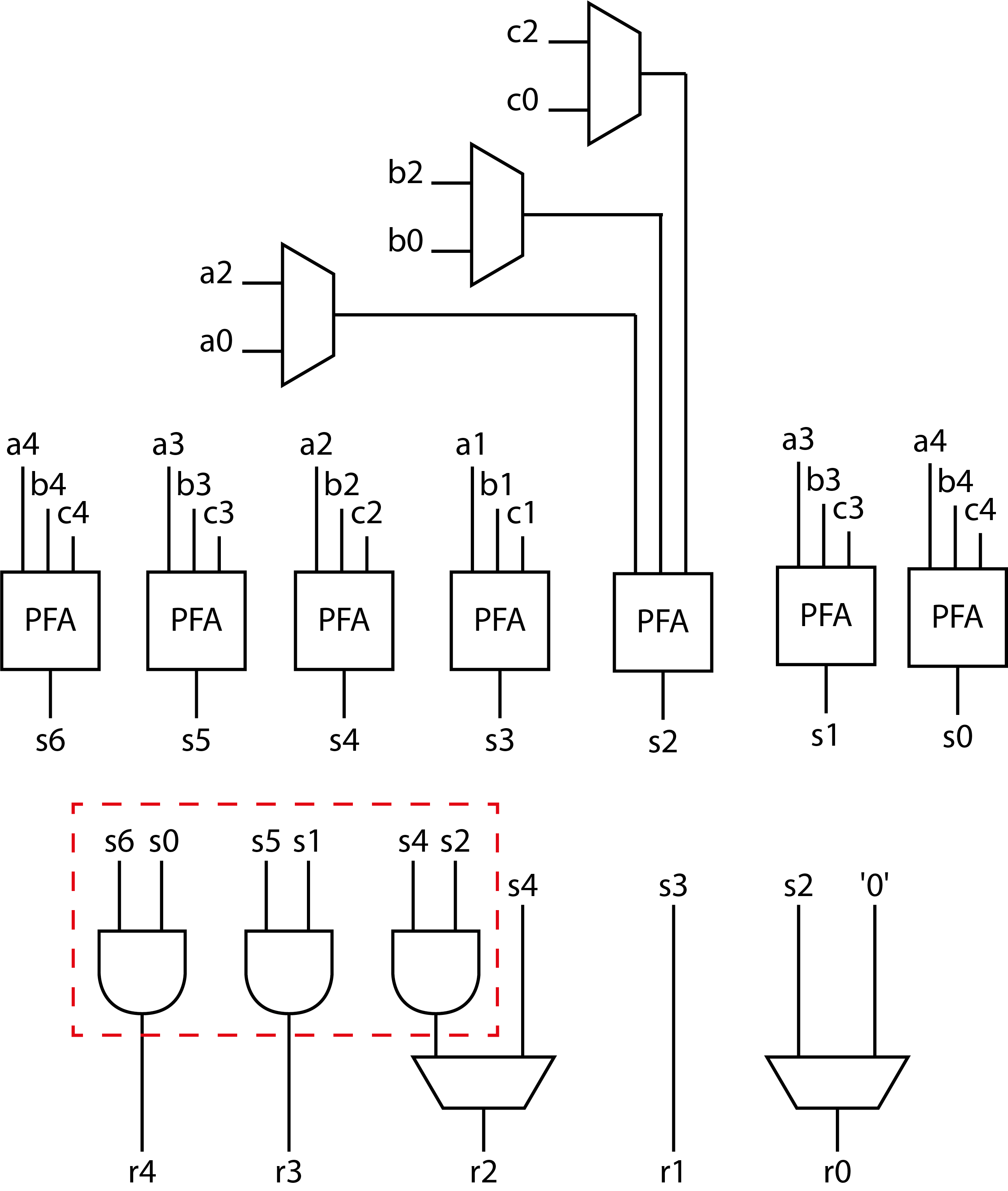}
    \caption{Adaptive adder architecture: $a$ and $b$ are inputs, $c$ is carry values and PFA stands for partial full adder}
    \label{fig:adapt_adder}
\end{figure}

The proposed architecture of the multiplier is presented in Fig. \ref{fig:mult} (the contributions and extensions to the logarithmic Mitchell multiplier are marked with red color). Assuming each operand has $n = 8$ bits, a Leading One Detector (LOD) circuit is used to find the index of the first ‘1’ bit in each operand. This index denoted as $k$, is the characteristic or integer part of the logarithm and has $log_2(n) = 3$ bits. The multiplier shifts the operands left by $k$ bits, aligning the leading one with the Most Significant Bit (MSB). $(n – 1)$ bits after the leading one represent the mantissa part denoted as $m$. The mantissa is truncated to $t = 5$ bits. The truncated operands are passed to the adaptive $(n – 1)$-bit adder that adds mantissa together and duplicates the addition of 2 or 3 MSBs depending on the $k$ value of the biggest operand for fault \textit{detection} and \textit{mitigation}. The architecture of the adaptive adder is shown in Fig. \ref{fig:adapt_adder}. The adder is based on the carry lookahead adder. Duplicated results are compared, and if there is a fault, the faulty bit is set to zero using AND gates (marked on the figure with a red rectangle). Due to the truncation of the mantissa, up to 2 Least Significant Bits (LSB) are excluded from the calculation, which affects only the bigger numbers with the $k$ equal to 7 or 6. This introduces a small error compared to the original Mitchell algorithm that is discussed in the results section. The $k$ values of the operands are added separately. This adder is replicated three times, as the order of the final output depends on the result of this addition. A majority voter selects the final result. Then, the antilogarithm algorithm is used to get the product of multiplication. The sum of $k$ values determines the position of the leading one in the output product, which is followed by the sum of the mantissa parts using the appropriate shift operation. 

\subsection{Adaptive adder}\label{adapt_add}
\begin{figure}[h]
    \centering
    \includegraphics[width = 0.25\textwidth]{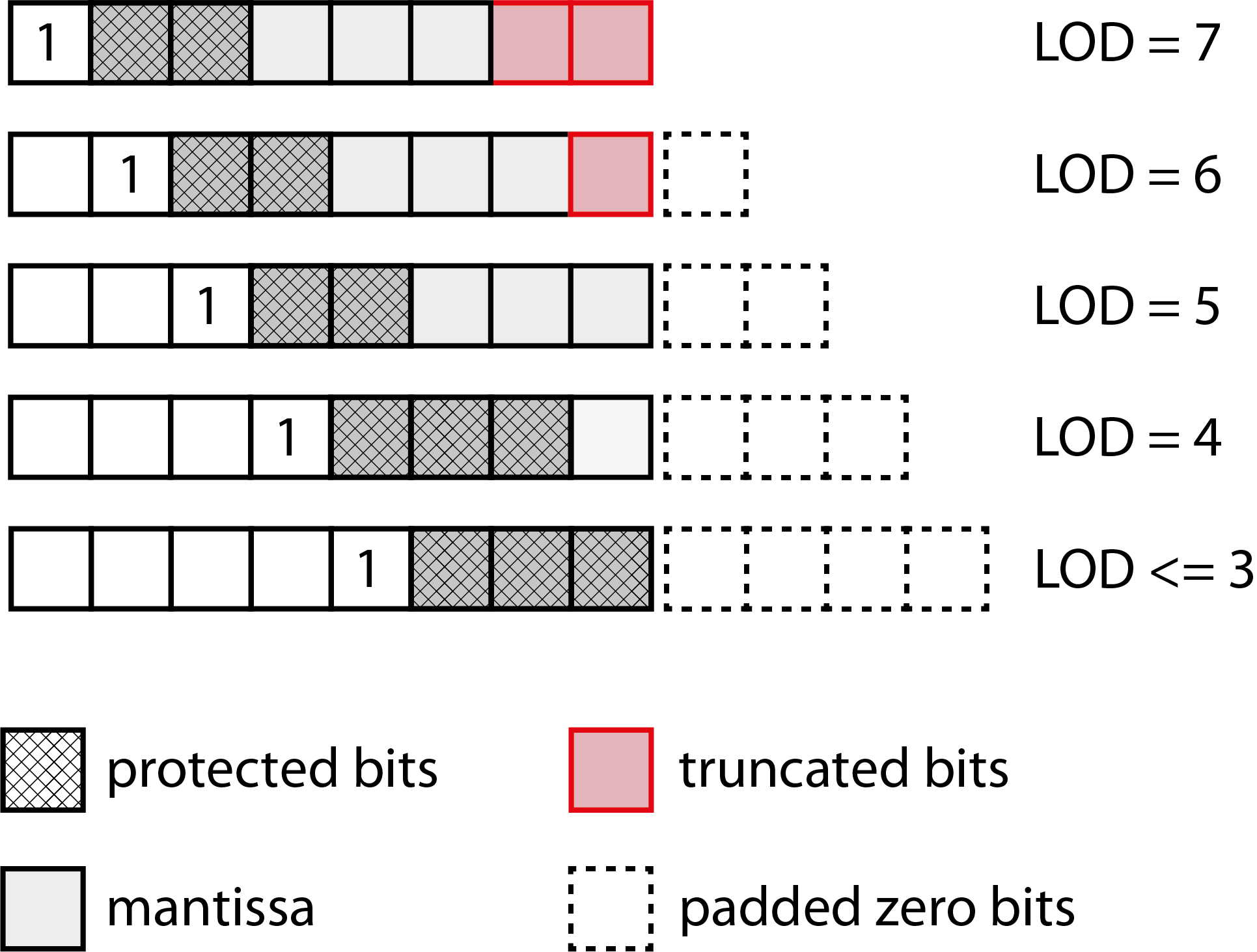}
    \caption{Fault-tolerance and error introduced based on different LOD cases}
    \label{fig:lod_cases}
\end{figure}

The adaptive adder is designed to perform fault detection and mitigation based on the LOD values of the multiplier inputs.
Fig. \ref{fig:lod_cases} shows the scheme in which the proposed multiplier introduces fault tolerance. As shown in this figure, five cases are considered. If the maximum LOD of the inputs is 7, two LSBs are discarded, and a two-bit adder of the adaptive adder is dedicated to recomputing the addition of two MSBs. These results are compared, and the mismatched bits are replaced by zeros in case of a mismatch.

For LOD~=~6, only the LSB is discarded, and two higher-order bits are protected the same way as in the previous case. When LOD~=~5, no bits are discarded, and two higher-order bits are protected. For LOD~=~4, three higher-order bits are protected, and only LSB is not monitored. In the case of LOD~$\le$~3, all bits are protected, enabling the proposed multiplier to provide comprehensive fault detection and mitigation for all inputs.

\section{Experimental Results}

\subsection{Experimental Setup}

In this paper, the FreePDK 45 nm Nangate technology library is used in Cadence Genus 2023 to compare the hardware characteristics of the proposed methods with the state-of-the-art. 
The impact on the accuracy of the proposed adaptive multiplier is studied on different networks (i.e., LeNet-5, AlexNet, and VGG-16) trained on MNIST and CIFAR-10 using 8-bit INT with the help of the ADAPT framework \cite{danopoulos2022adapt}. Finally, the impact of the proposed multiplier on the reliability of DNNs is studied using AlexNet and VGG-16. To perform the reliability simulations for the case studies, a systolic-array simulator is developed and integrated into the ADAPT framework, and the impact of transient faults in the multiply-accumulate (MAC) units of the systolic array is studied in the network.

\emph{Random fault injection.} 
Fault injection is performed, assuming the single bit-flip faults in the network's MAC operation of a systolic array for reliability assessment. Considering a prohibitively large number of fault combinations required for the multiple-bit fault model, it has been shown that high fault coverage obtained using the single-bit model results in a high fault coverage of multiple-bit faults \cite{bushnell2004essentials}. 
According to the adopted single-bit fault model, a random bit-flip is injected into a random MAC unit of the systolic array core at a random execution time of the network, and the whole test set is fed to the network to obtain the accuracy of the network. This process is repeated several times to reach an acceptable confidence level, based on \cite{leveugle2009statistical}. This work provides an equation to reach 95\% confidence level and 1\% error margin.

\subsection{Hardware utilization}
In this section, the adaptive multiplier is compared in terms of power and area with state-of-the-art designs.

In Table \ref{mult_r}, the accuracy, efficiency, and fault tolerance (FT) of 8-bit approximate multipliers are compared with the proposed method. Wallace, DRUM\cite{hashemi2015drum}, TOSAM\cite{vahdat2019tosam}, and ScaleTrim\cite{farahmand2023scaletrim} are used for this comparison. The accuracy is reported using Mean Absolute Relative Error (MARE).
The proposed multiplier has similar hardware parameters to the state-of-the-art approximate multipliers with similar accuracy while providing reliability improvement with fault detection and mitigation capability.

\begin{table}[h]
\caption{Accuracy and efficiency of 8-bit approximate multipliers compared with the proposed method}
\label{mult_r}
\begin{center}

\begin{tabular}{|c|c|c|c|c|c|c|}
\hline
\begin{tabular}[c]{@{}c@{}}Multiplier\\ Architecture\end{tabular} &
\begin{tabular}[c]{@{}c@{}}Delay\\ (ns)\end{tabular} &
\begin{tabular}[c]{@{}c@{}}Power\\ ($\mu$W)\end{tabular} &
\begin{tabular}[c]{@{}c@{}}Area\\ ($\mu$$m^2$)\end{tabular} &
\begin{tabular}[c]{@{}c@{}}MARE\\ (\%)\end{tabular} & \multicolumn{1}{l|}{FT} &
\begin{tabular}[c]{@{}c@{}}PDP\\ (pJ)\end{tabular}\\ \hline
\cellcolor{gray!25}Exact (Wallace)      & \cellcolor{gray!25}0.85 & \cellcolor{gray!25}360 & \cellcolor{gray!25}417 & \cellcolor{gray!25}0.00 & \cellcolor{gray!25}No & \cellcolor{gray!25}306\\ \hline
DRUM(3)      & 0.70 & 104 & 143 & 12.6 & No & 72.8\\ \hline
TOSAM(0,3)   & 0.68 & 144 & 198 & 7.7  & No & 97.9\\ \hline
DRUM(4)      & 1.00 & 172 & 208 & 6.4  & No & 172\\ \hline
TOSAM(1,5)   & 0.88 & 231 & 291 & 4.1  & No & 203.2\\ \hline
ScaleTrim(4,8) &   1.8 & 143 & 216 & 3.3    & No &  257.4\\ \hline
\cellcolor{green!25}AdAM     &   \cellcolor{green!25}1.13   & \cellcolor{green!25}165 & \cellcolor{green!25}152 & \cellcolor{green!25}4.7  & \cellcolor{green!25}Yes & \cellcolor{green!25}186.45
\\ \hline
\end{tabular}
\end{center}
\end{table}

\subsection{DNN accuracy}
Table~\ref{acc} compares the accuracy of different CNN architectures using the proposed approximate multiplier with the baseline accuracy using the exact multiplier. The evaluation shows that the accuracy of DNN with the proposed method is very close to the baseline. Hence, the proposed multiplier has a negligible effect on the accuracy of DNNs.

\begin{table}[h]
\caption{Accuracy comparison of different CNNs with an exact (baseline) and the proposed approximate multiplier}
\label{acc}
\begin{center}
\begin{tabular}{|c|c|c|}
\hline
DNN & \begin{tabular}[c]{@{}c@{}}Baseline\\ accuracy (\%)\end{tabular} & \begin{tabular}[c]{@{}c@{}}With proposed\\ multiplier (\%)\end{tabular} \\ \hline
\begin{tabular}[c]{@{}c@{}}LeNet-5 (MNIST)\end{tabular}    & 93.8 & 94.1 \\ \hline
\begin{tabular}[c]{@{}c@{}}AlexNet (CIFAR-10)\end{tabular} & 78.0 & 77.7 \\ \hline
\begin{tabular}[c]{@{}c@{}}VGG-16 (CIFAR-10)\end{tabular} & 93.4 & 94.0 \\ \hline
\end{tabular}
\end{center}
\end{table}

\subsection{Reliability analysis}
To showcase the impact of the AdAM multiplier on reliability, the fault injection simulations are performed on AlexNet and VGG-16 with four different configurations. The DNN reliability is evaluated by comparing the output probability vector of the golden run (i.e. the DNN that behaves as expected, without faults) and the faulty run (i.e. the DNN that includes the fault). 
The SDC rate is defined as the proportion of faults that caused misclassification in comparison with the golden model. Since in DNNs, there is often not a single correct output, but a list of ranked outputs, each with a confidence score, the new criteria to determine what constitutes an SDC for a DNN application is defined in  \cite{9926241}. 

Fig. \ref{re} demonstrates the fault tolerance comparison and reliability improvement of different networks by using the exact unprotected multipliers, using approximate unprotected multipliers (ScaleTRIM), using exact multipliers protected with TMR, and using AdAM. As illustrated, TMR has 100\% of protection but it also requires about 200\% of area overhead. Despite using TMR in our architecture for a small adder, we introduce very high reliability improvement without introducing hardware overhead. Since the main objective of the proposed multiplier is to have the best trade-off between PDP and vulnerability (accuracy drop due to the fault), Fig.~\ref{pdp_vul} illustrates this comparison. In these charts, the closer to the origins (0,0), the higher the cost-efficiency of the fault tolerance, i.e. lower vulnerability and PDP. As shown, TMR is an inefficient solution for edge AI applications because of its high PDP, while the proposed method (AdAM) is the closest to the origin. 
\begin{figure}[h]
    \centering
    \includegraphics[width=1\linewidth]{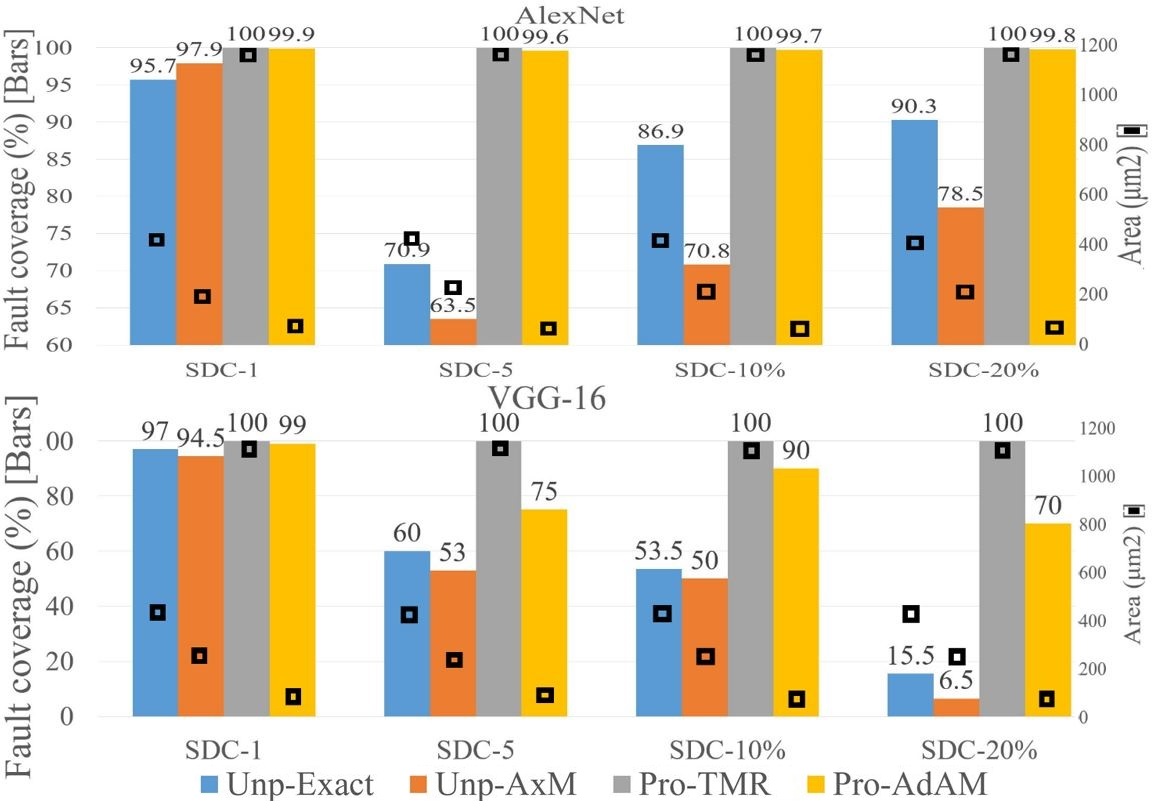}
    \caption{Hardware efficiency (area) and fault resilience (fault coverage) trade-offs in AlexNet (up) and VGG-16 (down). Unp-exact: unprotected exact multiplier, Unp-AxM: unprotected approximate multiplier, Pro-TMR: exact multiplier protected by TMR, Pro-AdAM: proposed multiplier}
    \label{re}
\end{figure}
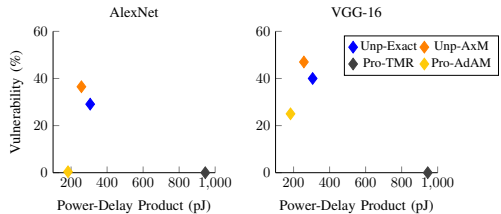
\begin{figure}[h!]
\centering
\begin{subfigure}{0.18\textwidth}
\begin{tikzpicture}[scale=0.55]
\begin{axis} [axis lines*=left, height=5cm, width=5.5cm, ymin=0, ymax=60, xmin=100, xmax=1000, 
              ytick={0,20,40,60,80,100},
              only marks, scatter, mark size=4pt, scatter src=explicit symbolic, scatter/classes={
              Diamond={mark=diamond*}},
              title=AlexNet,
              ylabel={Vulnerability (\%)}, xlabel={Power-Delay Product (pJ)}, ylabel near ticks]
\addplot [color=blue] table [meta=label] {
x    y   label
306 29.1 Diamond
};
\addplot [color=orange] table [meta=label] {
x    y   label
257.4 36.5 Diamond
};
\addplot [color=darkgray] table [meta=label] {
x    y   label
945 0 Diamond
};
\addplot [color=yellow!60!orange] table [meta=label] {
x    y   label
183.06 0.4 Diamond
};
\end{axis}
\end{tikzpicture}
\end{subfigure}%
\begin{subfigure}{0.18\textwidth}
\begin{tikzpicture}[scale=0.55]
\begin{axis} [axis lines*=left, height=5cm, width=5.5cm, ymin=0, ymax=60, xmin=100, xmax=1000, 
              ytick={0,20,40,60,80,100},
              only marks, scatter, mark size=4pt, scatter src=explicit symbolic, scatter/classes={
              Diamond={mark=diamond*}},
              title=VGG-16,
              xlabel={Power-Delay Product (pJ)}, ylabel near ticks,
              legend entries={Unp-Exact,Unp-AxM, Pro-TMR, Pro-AdAM},
              legend style={font=\small,at={(0.9,0.97)},anchor=north,legend columns=2}]
\addplot [color=blue] table [meta=label] {
x    y   label
306 40 Diamond
};
\addplot [color=orange] table [meta=label] {
x    y   label
257.4 47 Diamond
};
\addplot [color=darkgray] table [meta=label] {
x    y   label
945 0 Diamond
};
\addplot [color=yellow!60!orange] table [meta=label] {
x    y   label
183.06 25 Diamond
};
\end{axis}
\end{tikzpicture}
\end{subfigure}%
\caption{PDP and vulnerability tradeoffs (considering SDC-5) in different methods}
\label{pdp_vul}
\end{figure}

\section{Conclusion}
In this paper, we propose an architecture of a novel adaptive fault-tolerant approximate multiplier tailored for ASIC-based DNN accelerators. AdAM employs an adaptive adder relying on an unconventional use of the leading one position value of the inputs for fault detection through the optimization of unutilized adder resources. The proposed architecture uses a lightweight fault mitigation technique that sets the detected faulty bits to zero. 
It is demonstrated that the proposed multiplier provides a reliability level close to the multipliers protected by Triple Modular Redundancy (TMR) while utilizing 63.54\% less area and having 39.06\% lower power-delay product compared to the exact multiplier.
\section{Acknowledgement}
This work was supported in part by the Estonian Research Council grant PUT PRG1467 "CRASHLESS“ and by Estonian-French PARROT project "EnTrustED".

\bibliographystyle{IEEEtran}
\bibliography{ref}

\begin{thebibliography}{10}
\providecommand{\url}[1]{#1}
\csname url@samestyle\endcsname
\providecommand{\newblock}{\relax}
\providecommand{\bibinfo}[2]{#2}
\providecommand{\BIBentrySTDinterwordspacing}{\spaceskip=0pt\relax}
\providecommand{\BIBentryALTinterwordstretchfactor}{4}
\providecommand{\BIBentryALTinterwordspacing}{\spaceskip=\fontdimen2\font plus
\BIBentryALTinterwordstretchfactor\fontdimen3\font minus \fontdimen4\font\relax}
\providecommand{\BIBforeignlanguage}[2]{{%
\expandafter\ifx\csname l@#1\endcsname\relax
\typeout{** WARNING: IEEEtran.bst: No hyphenation pattern has been}%
\typeout{** loaded for the language `#1'. Using the pattern for}%
\typeout{** the default language instead.}%
\else
\language=\csname l@#1\endcsname
\fi
#2}}
\providecommand{\BIBdecl}{\relax}
\BIBdecl

\bibitem{ahmadilivani2023systematic}
M.~H. Ahmadilivani and et~al, ``A systematic literature review on hardware reliability assessment methods for deep neural networks,'' \emph{arXiv preprint arXiv:2305.05750}, 2023.

\bibitem{taheri2022dnn}
M.~Taheri, ``Dnn hardware reliability assessment and enhancement,'' \emph{27th IEEE European Test Symposium (ETS).}, May 2022.

\bibitem{mahatme2011comparison}
N.~Mahatme and et~al, ``Comparison of combinational and sequential error rates for a deep submicron process,'' \emph{TNS}, pp. 2719--2725, 2011.

\bibitem{taheri2022novel}
M.~Taheri and et~al, ``A novel fault-tolerant logic style with self-checking capability,'' in \emph{IOLTS}.\hskip 1em plus 0.5em minus 0.4em\relax IEEE, 2022, pp. 1--6.

\bibitem{farahmand2023scaletrim}
E.~Farahmand and et~al, ``{ScaleTRIM:} scalable truncation-based integer approximate multiplier with linearization and compensation,'' \emph{arXiv preprint arXiv:2303.02495}, 2023.

\bibitem{Henkel}
G.~Armeniakos and et~al, ``Hardware approximate techniques for deep neural network accelerators: A survey,'' \emph{ACM Comput. Surv.}, vol.~55, no.~4, nov 2022.

\bibitem{Guanpeng}
G.~Li and et~al, ``Understanding error propagation in deep learning neural network {(DNN)} accelerators and applications,'' in \emph{SC'17}, 2017.

\bibitem{10140043}
M.~H. Ahmadilivani and et~al, ``Special session: Approximation and fault resiliency of {DNN} accelerators,'' in \emph{VTS}, 2023, pp. 1--10.

\bibitem{taheri2023appraiser}
M.~Taheri and et~al, ``Appraiser: Dnn fault resilience analysis employing approximation errors,'' in \emph{DDECS}.\hskip 1em plus 0.5em minus 0.4em\relax IEEE, 2023, pp. 124--127.

\bibitem{taheri2024exploration}
M.~Taheri, N.~Cherezova, and et~al, ``Exploration of activation fault reliability in quantized systolic array-based dnn accelerators,'' \emph{arXiv preprint arXiv:2401.09509}, 2024.

\bibitem{taheri2023deepaxe}
M.~Taheri, M.~H. Ahmadilivani, and et~al, ``Deepaxe: A framework for exploration of approximation and reliability trade-offs in {DNN} accelerators,'' in \emph{ISQED}.\hskip 1em plus 0.5em minus 0.4em\relax IEEE, 2023, pp. 1--8.

\bibitem{mittal2020survey}
S.~Mittal, ``A survey on modeling and improving reliability of {DNN} algorithms and accelerators,'' \emph{J. Syst. Archit.}, vol. 104, p. 101689, 2020.

\bibitem{ammes2023atmr}
G.~Ammes and et~al, ``{ATMR} design by construction based on two-level {ALS},'' in \emph{36th SBCCI}.\hskip 1em plus 0.5em minus 0.4em\relax IEEE, 2023, pp. 1--6.

\bibitem{arifeen2020approximate}
T.~Arifeen and et~al, ``Approximate triple modular redundancy: A survey,'' \emph{IEEE Access}, vol.~8, pp. 139\,851--139\,867, 2020.

\bibitem{Mitchell}
J.~N. Mitchell, ``Computer multiplication and division using binary logarithms,'' \emph{IRE Trans. on Electronic Computers}, pp. 512--517, 1962.

\bibitem{danopoulos2022adapt}
D.~Danopoulos and et~al, ``Adapt: Fast emulation of approximate {DNN} accelerators in {Pytorch},'' \emph{IEEE TCAD}, 2022.

\bibitem{bushnell2004essentials}
M.~Bushnell and V.~Agrawal, \emph{Essentials of electronic testing for digital, memory and mixed-signal {VLSI} circuits}.\hskip 1em plus 0.5em minus 0.4em\relax SSBM, 2004, vol.~17.

\bibitem{leveugle2009statistical}
R.~Leveugle and et~al, ``Statistical fault injection: Quantified error and confidence,'' in \emph{DATE}, 2009, pp. 502--506.

\bibitem{hashemi2015drum}
S.~Hashemi, R.~I. Bahar, and S.~Reda, ``Drum: A dynamic range unbiased multiplier for approximate applications,'' in \emph{ICCAD}, 2015, pp. 418--425.

\bibitem{vahdat2019tosam}
S.~Vahdat and et~al, ``{TOSAM:} an energy-efficient truncation-and rounding-based scalable approximate multiplier,'' \emph{TVLSI}, vol.~27, no.~5, pp. 1161--1173, 2019.

\bibitem{9926241}
G.~Li, S.~K.~S. Hari, M.~Sullivan, T.~Tsai, K.~Pattabiraman, J.~Emer, and S.~W. Keckler, ``Understanding error propagation in deep learning neural network (dnn) accelerators and applications,'' in \emph{SC17}, 2017.

\end{thebibliography}

\end{document}